\documentclass{article} %
\usepackage{iclr2023_conference,times}

\usepackage{amsmath,amsfonts,bm}

\def\eqref#1{equation~\ref{#1}}

\def\1{\bm{1}}

\DeclareMathAlphabet{\mathsfit}{\encodingdefault}{\sfdefault}{m}{sl}
\SetMathAlphabet{\mathsfit}{bold}{\encodingdefault}{\sfdefault}{bx}{n}

\usepackage{graphicx}
\usepackage{svg}
\usepackage{relsize}
\usepackage[hidelinks]{hyperref}
\usepackage{url}
\usepackage{multirow}
\usepackage[stable]{footmisc}
\usepackage{pifont}
\usepackage[all, defaultlines=3]{nowidow}

\usepackage[disable]{todonotes}

\title{Memory Efficient Mixed Precision Optimizers}

\author{Basile Lewandowski\thanks{Corresponding author. Work carried out as a Master’s thesis from Sorbonne University’s computer science degree, supervised by Atli Kosson}\,, Atli Kosson \\
Machine Learning Optimization laboratory\\
École Polytechnique Fédérale de Lausanne\\
\texttt{firstname.lastname@epfl.ch}
}

\iclrfinalcopy %
\begin{document}

\maketitle

\begin{abstract}
Traditional optimization methods rely on the use of single-precision floating point arithmetic, which can be costly in terms of memory size and computing power. However, mixed precision optimization techniques leverage the use of both single and half-precision floating point arithmetic to reduce memory requirements while maintaining model accuracy. We provide here an algorithm to further reduce memory usage during the training of a model by getting rid of the floating point copy of the parameters, virtually keeping only half-precision numbers. We also explore the benefits of getting rid of the gradient's value by executing the optimizer step during the back-propagation. In practice, we achieve up to 25\% lower peak memory use and 15\% faster training while maintaining the same level of accuracy.
\end{abstract}

\section{Introduction}
The global trend in machine learning networks is that larger models yield more accurate results. Consequently, new models have been designed with an ever-increasing number of parameters. However, training such models can be consuming in term of computing power, which is why recent work has been aimed towards alternatives to the single-precision arithmetic (fp32). Most recent models are so large they cannot fit on a single GPU using a traditional training framework: LLaMA (Meta AI's latest large language model) would require for instance 560 GB of memory, which is far more than state of the art GPU can offer. 

As neural networks grow larger and larger, the need to reduce their memory footprint has become increasingly imperative. While prior research has focused on increasing speed, there remain room for improvements in the way GPU memory is used. Indeed, the common approach to maintaining some accuracy on the parameters while training on a half precision (fp16) model is to keep a master copy of them in full floating-point precision. The drawback of doing so is that every parameter now has to be saved in memory in both fp16 and fp32, further expending the charge for GPU memory.

Typical mixed precision training uses half precision values during the forward pass and single precision for the parameters' update. For each parameter, the components stored on GPU memory are then: the single precision value of the model weight and its half precision copy (6 bytes), the optimizer state (dependent on the optimizer), the gradient (usually 4 bytes), and some additional values like the forward activations whose size may vary depending on the model.
To lower the memory pressure, solutions have already been developed towards smaller optimizer footprint. Indeed, memory requirements for modern model training are often dictated by the optimizer state, with up to twice as much memory required for each parameter. Alternative optimizers, such as Adafactor or 8bit-Adam, already offer a remedy to this problem by changing the way state memory is stored. Where a standard Adam optimizer would require 8 bytes of state memory by parameter, these optimizers respectively uses 4 bytes and 2 bytes only. The model's parameters and their gradients then becomes the main challenge to enhance memory use.

Our work aims at removing the additional memory cost incurred by the fp32 copy of the parameter by keeping in memory only the difference between the original parameter and its fp16 value. We also get rid of the gradient value by directly applying the update during the backward pass, thus relieving the need to keep those values at all time. For a given parameter, this leads to - at least - 6 bytes less having to remain stored on the memory.

Our method does not necessitate any alterations to the hyperparameters or the training framework. It is designed to fit models that require an extensive amount of memory for their training, such as Large Language Models, Image Classification or Generation, Natural Language Processing, Object Detection.

\begin{table}[b]
    \centering
    \caption{Classical floating point formats}
    \begin{tabular}{c c c c c}
        \textbf{Name} &  \textbf{Length} & \textbf{Sign} & \textbf{Exponent length} & \textbf{Significand length} \\
        \hline
        Float & 32 bits & 1 bit & 8 bits & 1 + 23 bits \\
        BFloat16 & 16 bits & 1 bit & 8 bits & 1 + 7 bits \\
        Half & 16 bits & 1 bit & 5 bits & 1 + 10 bits 
    \end{tabular}
    \label{tab1}
\end{table}

\todo[inline]{Atli: Here I would give a high level overview of standard mixed precision and then point out the memory redundancy that we aim to fix. See link in comment %
for a similar overview in terms of number of bytes required for each parameter.}

\subsection{Floating Point Format Basics}
We remind here the basic principles of floating point arithmetic to frame our work thereafter. The representation of common floating point numbers is ruled by the IEEE-754 Standard \cite{4610935}. According to it, a binary floating point number $x$ is represented in memory by :
\begin{itemize}
    \item its \textbf{sign} $s_x$ (0 if positive, 1 if negative)
    \item its \textbf{exponent} $e_x$, an integer representing the order of magnitude of $x$
    \item its \textbf{mantissa (significand)} $m_x$, representing its decimal values w.r.t. $e_x$. The first non-zero bit is considered implicit and is not actually stored in memory.
\end{itemize}
such that $x=(-1)^{s_x}*m_x*2^{e_x}$

This representation is somewhat analog to the standard scientific notation (for instance $\pi=+3.14*10^0$). The encoding used for $e_x$ and $m_x$ depends on which format is used, as described in \autoref{tab1}. 

The exponent value is represented with a bias so that numbers can be ordered lexicographically. Some exponent values are reserved for special numbers : $e_{max}$ (maximum value) for infinities and \textsf{NaN}, zero for subnormal numbers (that do not have an implicit bit), including zero itself. Some compilers (eg. \textsf{fast-math} option) or non-standard formats (like \textsf{bfloat16}) do not use the subnormal representation. 

\section{Related Work}

The standard framework for mixed precision training on neural networks is the one described by \cite{micikevicius_mixed_2018}. Especially, this work describes the use of a full precision copy for each parameter to prevent the training from failing because weight updates are too small compared to weight values. It also proposes the use of loss scaling and full precision arithmetic in some cases where it is necessary to prevent the gradient from instabilities.

Alternatives solutions have considered the use of smaller floating point format (16bits and less), with different ways of maintaining some level of accuracy. One common method to ensure the values used remain accurate enough to allow for an efficient training is quantization : splitting the data in several chunks and representing each number with only a few bits to describe their value on a scale from the minimum of the chunk to its maximum. This allow for significant decrease in the memory footprint of the optimizer states (\cite{dettmers20228bit}) or for the gradients in a context of distributed training (\cite{alistarh2017qsgd}).  Quantization has also been extensively considered for inference-side computations (\cite{gholami2021survey}), but the scope of our work is the improvement of the training phase.

The use of Brain floating point format (bf16) is often preferred for 16bits only training, as it is more robust against gradient absorption in the updates and does not always necessitate some change in the hyper-parameters (\cite{pedram_bf16, kalamkar_study_2019}). 8bits floating point format can lead to promising results, and it is yet to determine which format is the best in a context of machine learning training (\cite{micikevicius_fp8_2022, wang_training_2018}). More extreme solutions have even considered the use of 1-bit binary networks (\cite{Qin_1bit}).

Another method proposed to train models on lower precision is the use of fixed point arithmetic. This solution represents floating values with a fixed number of fractional and integer bits. In this representation, a fixed number of bits are assigned to both parts of a decimal number. The position of the binary point determines the scaling and precision of the fixed-point value. This system may enable some interesting performance improvements but it is usually hard to maintain a sufficient accuracy.

\section{Implementation}
We have developed two methods to reduce the need for memory in the context of mixed precision training. The first approach aims at reducing the memory weight of the parameters by removing the full precision copy stored to maintain accuracy during the training. The second one is to get rid of the gradient values as soon as they are computed in order to avoid having all of the gradients stored on the GPU memory.
\subsection{16bits only Mixed-Precision}
As stated earlier, a classic implementation of mixed precision optimization typically stores in memory both a fp16 and a fp32 value, whereas our approach consists of storing only the difference between the two formats, thus resulting of at least a third less memory dedicated to the storage of a model's parameter. The stored part of the parameter is then used during its update in the backward pass.

Classic 16bits floating point format (fp16) contains 10 bits of significand whereas alternative Brain-floating point (bf16) contains only 7bits. There is therefore respectively 13 and 16 bits of precision to keep if we want to maintain a full fp32 accuracy. We also explore the performance when keeping only part of those bits.

To do so, we have developed an overload of the arithmetic operators used in the parameters' update (elementwise add, multiply, divide and their classic combinations) that performs the operation in full precision using the extra bits saved separately and outputs both the updated 16 bits float and its extra bits. We use for that a custom CUDA kernel where each thread handles the values stored in the same memory slot. Since there is no efficient way to access arbitrarily sized bit strings in the memory, the accesses are made through chunks of 32 bits (int32). Each GPU thread then handles the operation on several values, depending on how many different extra-stored values can fit on 32 bits.

\begin{figure}[tb]
    \centering
    \includegraphics[width=\columnwidth]{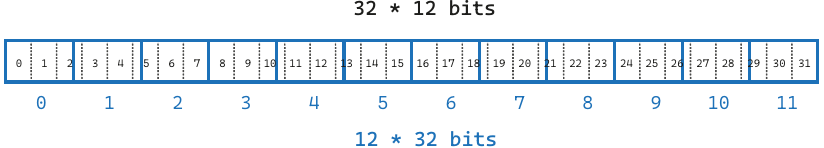}
    \caption{Storage example for 12 extra-bits: the extra storage is created and accessed using twelve 32-bits integers, the bits stored to keep some accuracy (12 bits per parameter) are distributed among 32 slots. With this solution, we ensure that extra-storage fits closely to the size of the parameter and that it is accessed efficiently through chunks of 32bits.}
    \label{fig1}
\end{figure}

To ensure using all of the memory when the bit-sizes are not multiple of 32, we operate the threads by groups corresponding to that bit-size, for instance values with 12 extra bits would require 12 threads working on 32 values stored in 12 int32 (see fig.~\ref{fig1}). The overlapping extra-bits (stored on two different int32) are modified using shared-memory to avoid data-race. 

\todo[inline]{I think we need a bit more clarification for the figure}

We provide an implementation for classic optimizers (Adam and SGD) to use transparently on a fp16 or bf16 model. Extra bits are stored by the optimizer and do not require any modification in the training framework to perform the weight updates. The extra precision provided ensures that small gradients are accurately represented, reducing the risk of gradient underflow and enabling successful model training.

This solution may incur some loss in accuracy on the 16bits floating point value. Indeed, saving only the first part of the 32bit significand is equivalent to applying a round-to-zero operation on the full precision value, which is known to be less accurate than the standard round-to-nearest used in this case. Even though the value stored is as precise as a classic fp32 training, the 16bits value used in the computations is then less accurate than a traditional mixed-precision framework. To overcome this issue, we have developed the option to use stochastic rounding when splitting the full precision value. To do so, we store one additional extra-bit to keep in memory whether or not the value was changed when rounded-up, so as to `un-round` the value before updating it.

Finally, we provide a Fused-version of the optimizers, since too many kernel launches can hinder performance on smaller parameters. The principle of fused optimizers is to reduce the amount of CUDA kernels launched during the optimization by handling the parameters as one only stream of values combined all together. That way, one can launch kernels on a larger number of values, so that we ensure the time needed for a kernel launch is amply covered by the time taken by the computations. However, since fused-optimizers access values independently of which parameter they belong to, they are for now only supported for 8 and 16 extra-bits.

\subsection{Fusing backward pass and optimizer step}
The classical implementation of the parameters update in a neural network is to compute the gradient for each parameter and then to use this gradient as input for an optimizer algorithm. The drawback of this system is that every gradient of every parameter has to be stored in the memory at all time (or at least as long as the optimizer has not stepped). This generates a consequent need for GPU memory as it nearly doubles the size of the parameters. Our solution to get rid of this pressure on the memory is to operate the optimization step as soon as the gradient is computed. 

In practice, we use PyTorch~(\cite{paszke_pytorch_2019}) automatic differentiation package to change the way gradients are computed during the backward pass, which enables us to update directly the parameters without saving any of their values. The optimization step is then performed by our backward pass function, which means every operations on the gradient (eg. clipping or scaling) has to be done through the optimizer. 
To accommodate every optimizer, our design requires that the optimizer's \verb|step| function is called for each parameters, which is unusual for a training framework. Our experiments shows that it is not a problem when using classic optimizers.
This solution prohibits any operation that would require every gradients of the model at the same time, but it is -- to our knowledge -- very uncommon. The case of gradient accumulation is discussed in \autoref{limitations}.

\section{Experimental results}

To validate the ability of our setup to maintain training performances comparable to a fp32 setup while using less memory, we carried out several model training from scratch. We consider the following models : the Deep Learning Recommendation Model (DLRM, \cite{naumov2019deep}), the Resnet-18 image classification model, T5 text-to-text transformer, DCGAN image generation.

\begin{table}[]
    \centering
    \begin{tabular}{c c c r}
    Task & \shortstack{Fused backward pass \\ (enabled \ding{51} or not \ding{56} )} & \shortstack{Peak memory \\ usage} & Time to complete \\ \hline\hline
    \multirow{2}{*}{MNLI} & \ding{56} & 29 930 MB & 10:41:28 \\
                          & \ding{51} & 26 818 MB & 11:07:34 \\ \hline
    \multirow{2}{*}{QNLI} & \ding{56} & 30 239 MB & 03:11:09 \\
                          & \ding{51}	& 26 826 MB & 03:16:48 \\ \hline
    \multirow{2}{*}{MRPC} & \ding{56} & 29 964 MB &    06:23 \\
                          & \ding{51}	& 26 649 MB &    06:40
    \end{tabular}
    \caption{Finetuning Performance on the GLUE Benchmark. We trained a large flan-T5 model on several tasks of the GLUE benchmark, comparing its performance when the backward pass is fused or not. Results show that peak memory usage is down by 11\%, while computation time increased only slightly. }
    \label{glue}
\end{table}

\begin{figure}[h]
    \centering
    \includeinkscape[width=.49\columnwidth]{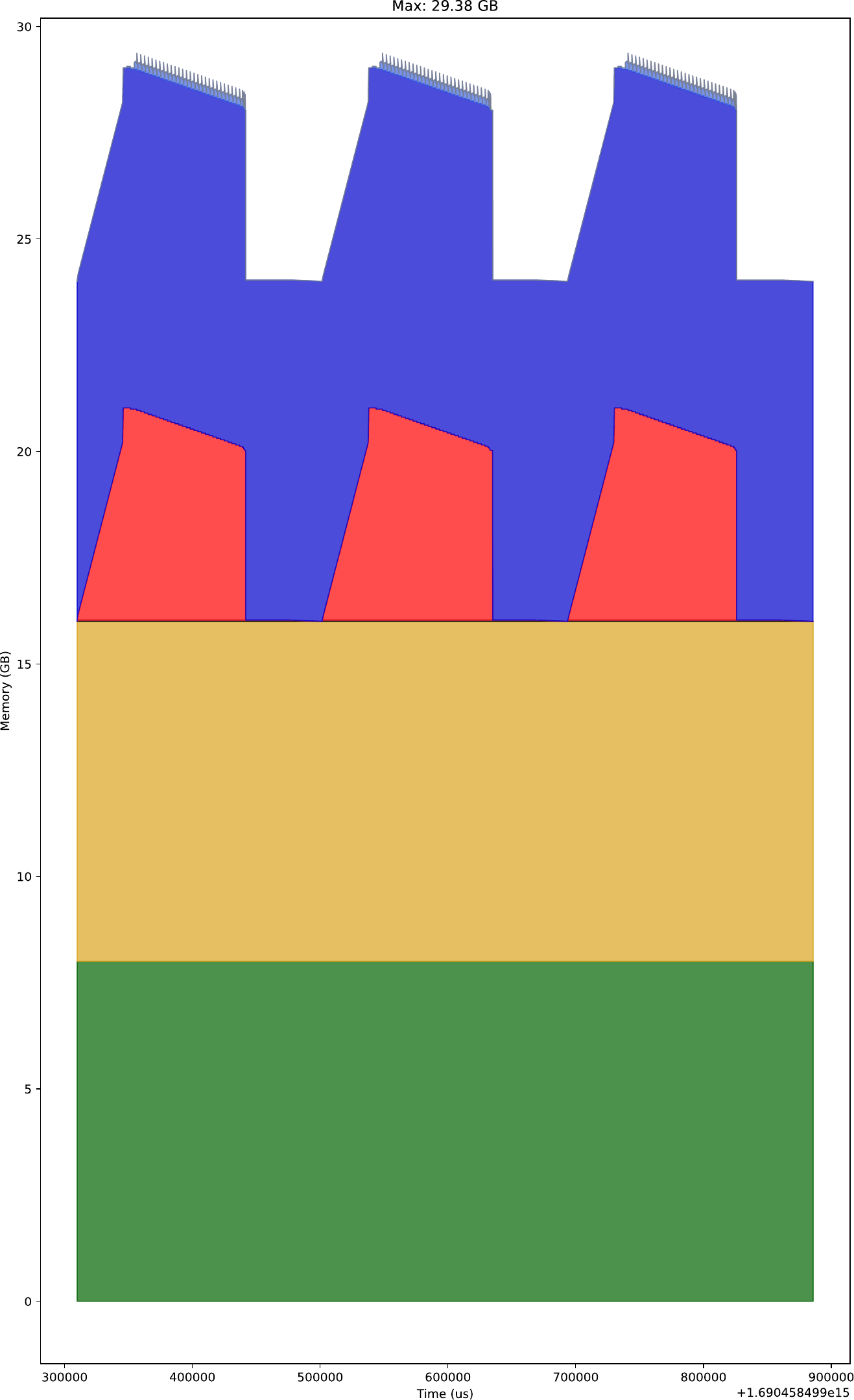}
    \includeinkscape[width=.49\columnwidth]{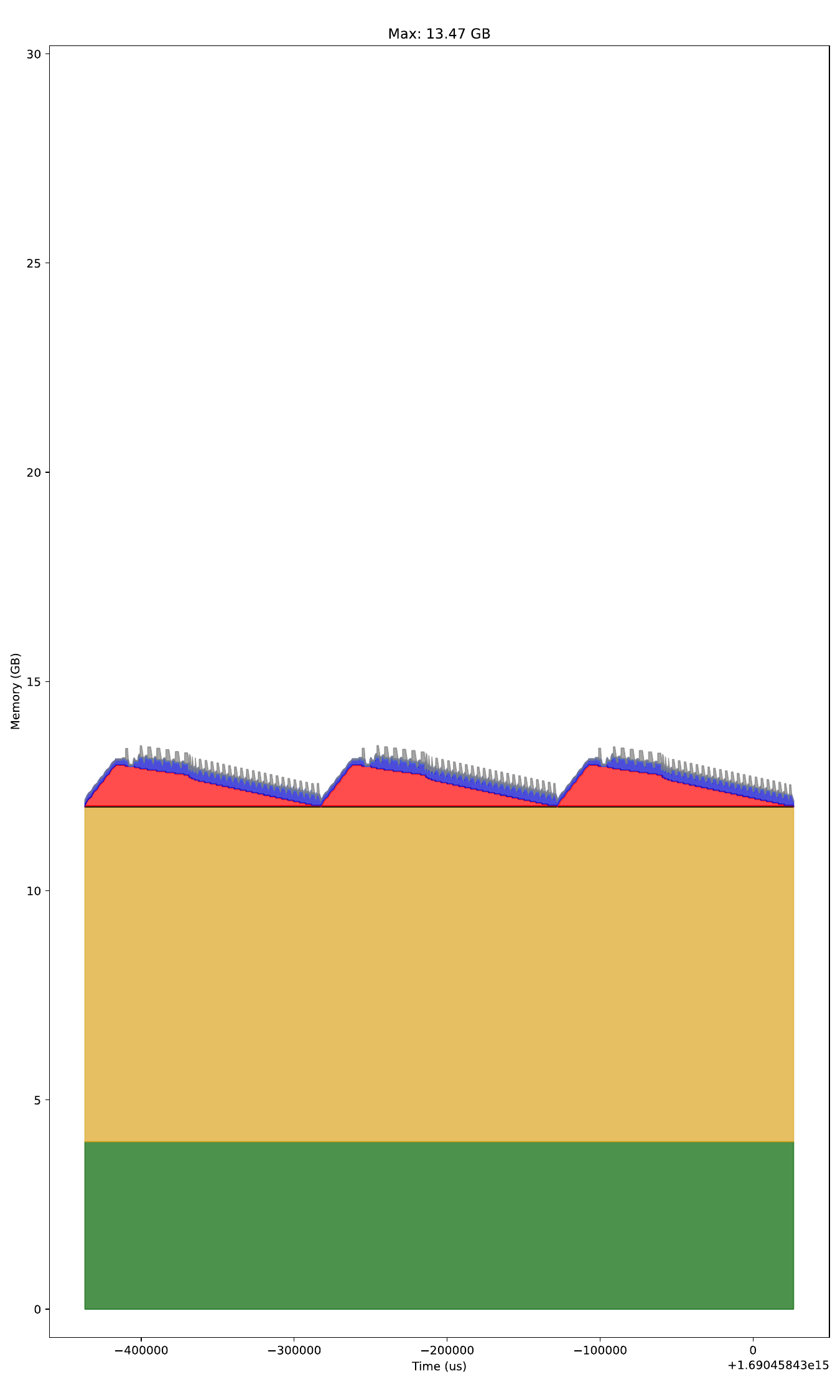}
    \includeinkscape[width=\columnwidth]{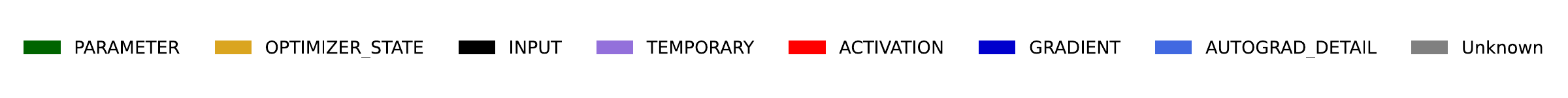}\hfill
    \caption{Memory Usage During A Synthetic Training. Left graph shows memory usage when using torch's standard mixed-precision framework, while the right one uses ours. Training consists of three steps on a dummy 2 billion parameters model using SGD with momentum optimizer. Our solution uses fp16 for the models parameters, the activation values and the gradients, and fp32 for the optimizer state. The extra-bits are considered as part of the optimizer state memory footprint. The profiler also considers the automatic mixed precision fp16 copy of the model's parameters as activation values, which is why they appear larger in the first figure. In total, peak memory usage is down by 54\% and could be further reduced by using an optimizer with reduced state memory. }
    \label{memuse}
\end{figure}

\paragraph{Memory Savings} Our solution enables for a smaller memory use in every phase of the training, and especially during the backward pass. The results displayed in \autoref{memuse} show we achieve up to 54\% lower peak memory use on a sample network compared to a standard mixed-precision training. This is close to the theoretical maximum loss of 60\% in this case. Fusing the back-propagation and the optimization provides on its own a 11\% decrease on peak memory while fine-tuning a T5 model on the GLUE benchmark, without significantly slowing down the training (detailed results are presented in \autoref{glue}). Overall, we achieve a reduction of 20 to 25\% in real training conditions, as it is the case with our ResNet18 experiment. We also observe that keeping only part of the mantissa (for instance 8bits on a fp16 training) can be enough to maintain the accuracy while further reducing the memory pressure (see fig.~\ref{dlrm} for instance).

\begin{table}[]
    \centering
    \begin{tabular}{*4c}
         Training framework & Execution Time & Accuracy (top 1) & Global GPU Memory usage  \\
         \hline
          fp32 & 42 min & 94.64 \% & 4 720 MB \\ 
          amp fp16 + fp32 & 24 min & 94.04 \% & 4 520 MB\\
          fp16+13 rstoc & 38 min & 94.10 \% & 3 868 MB\\
          fp16+8 & 20 min & 94.06 \% & 3 776 MB\\
          fp16 & 23 min & 93.79 \% & 3 840 MB \\
          fp8+8 & -- & 93.96 \% & -- \\
          fp8+4 & -- & 65.53 \% & -- \\
          fp8 & -- & 8.78 \% & --
    \end{tabular}
    \caption{Training performance on Resnet18 with CIFAR10. Our solution outperforms classic mixed-precision training both in terms of speed and memory use. This context shows too little difference in accuracy between the different formats to draw any conclusion. Training was performed on a Nvidia V100-SXM GPU.\\
    This hardware does not natively support fp8, therefore we emulated 8bits precision on fp16 values (e5m2 fp8 is equivalent to a truncated fp16), which is why execution time and memory usage is not relevant in that case. Moreover, fp8-only training could yield better results if it were part of a more adapted framework, it is displayed here only for comparison purposes.}
    \label{metrics}
\end{table}

\paragraph{Accuracy} Our experiments show our solution achieve the same accuracy levels as mixed precision. In most of the models we tested, accuracy obtained with 16bits formats is close to the results of full precision training. However, when 16bits training tends to diverge, our system prevent any massive loss of accuracy. On the recommendation model we trained, we match the accuracy of the full precision training by adding only 8bits of precision. As can be seen in fig.~\ref{dlrm}, this allows us to run fp16 training where it would diverge in a standard context.
In the case of ResNet18, our experiments showed no relevant difference in accuracy between the different format, which led us to experiment with 8bits floating point format (fp8). The results summarized in \autoref{metrics} indicates our mixed-precision framework could also provide a fp16 accuracy on a 8bits model. This format however lacks compatible hardware for now and is not yet supported by our solution.
On the DCGAN model, our optimizer produces significantly better results than a standard bf16 training, but it does not exactly match the results of full precision training. This is hard to evaluate performance difference in this case because the results are better or worse than the fp32 training depending on which metric we use. 

\begin{figure}
    \centering
    \includeinkscape[width=.5\textwidth]{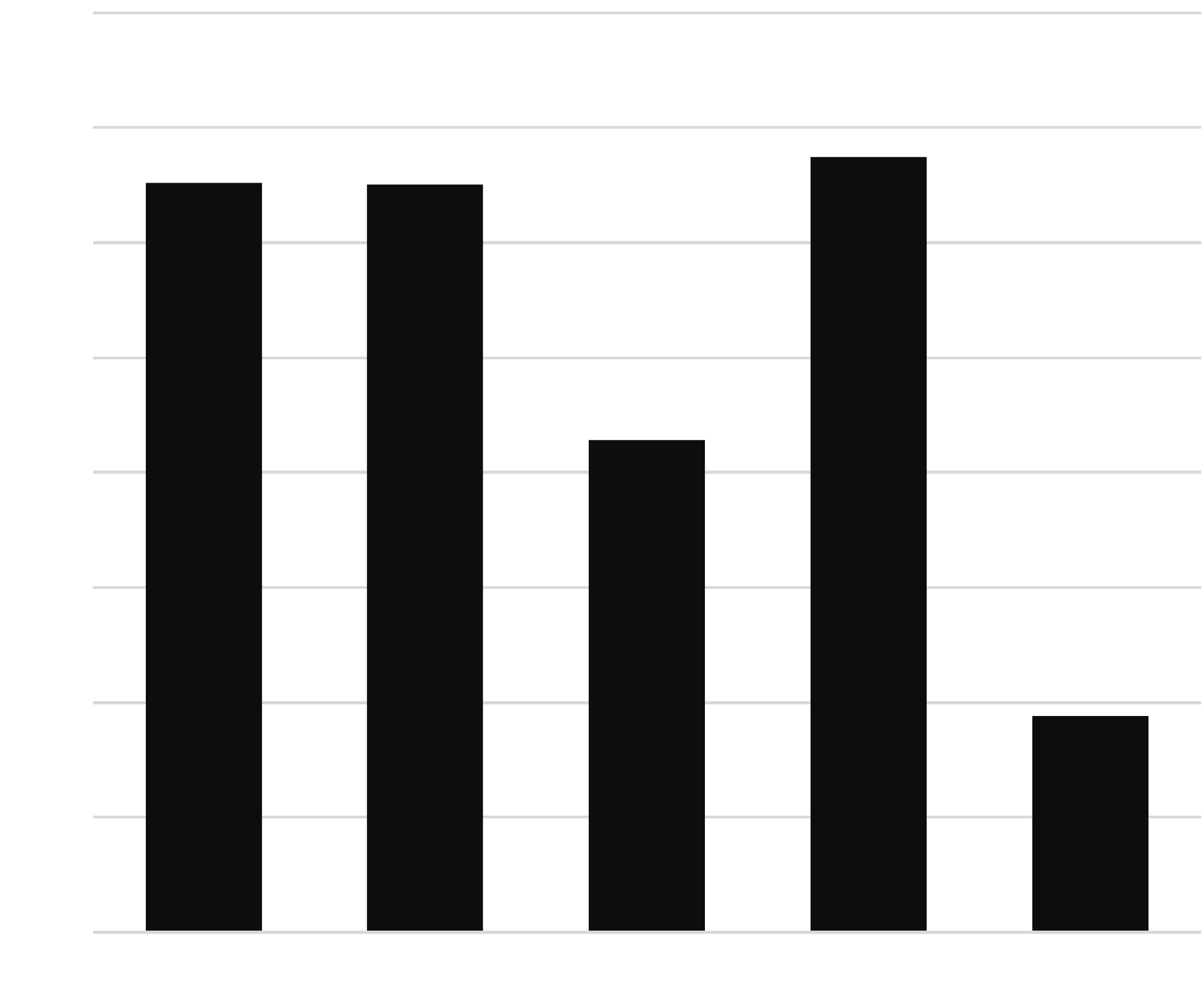}
    \caption{Throughput for an arithmetic operation on tensors. Data is shown as percentage of the theoretical GPU bandwidth. Experiments were run on a NVIDIA A100-40GB-SXM GPU}
    \label{throughput}
\end{figure}

\begin{figure}
    \centering
    \includeinkscape[scale=.25, pretex=\relscale{0.9}]{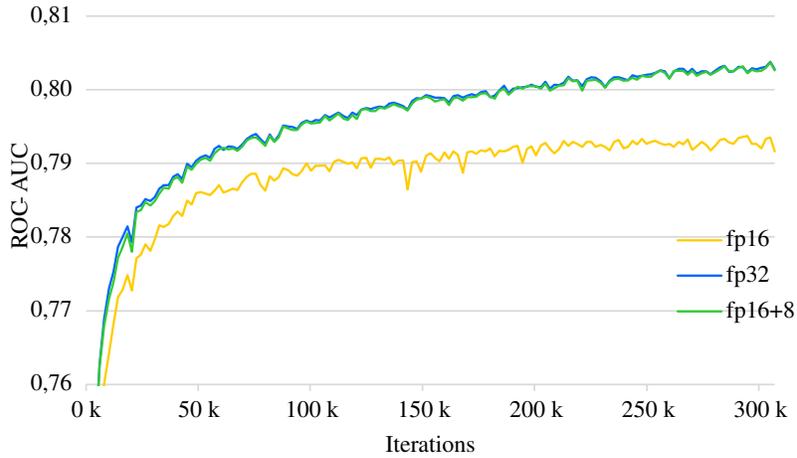}
    \caption{Accuracy improvement during training w.r.t classical fp16 framework on a DLRM model. In this case, 8 extra-bits of storage are sufficient to achieve the same accuracy as a standard single-precision setup. While a vanilla fp16 model leads to losing part of the model's accuracy, our optimizer enables us to avoid this loss.}
    \label{dlrm}
\end{figure}

\paragraph{Efficiency} The performance of our optimization techniques varies with the number of extra-bits we use. The absolute throughput of our operators shows that they are as efficient as torch's standard ones, and somewhat weaker for extra-bits sizes that rely heavily on shared memory (see \autoref{throughput}). On actual model training, our solution achieve up to 16\% faster training on the Resnet-18 model, as compared to classic mixed-precision training.

\begin{figure}
    \centering
    \includeinkscape[width=.5\textwidth, pretex=\relscale{0.5}]{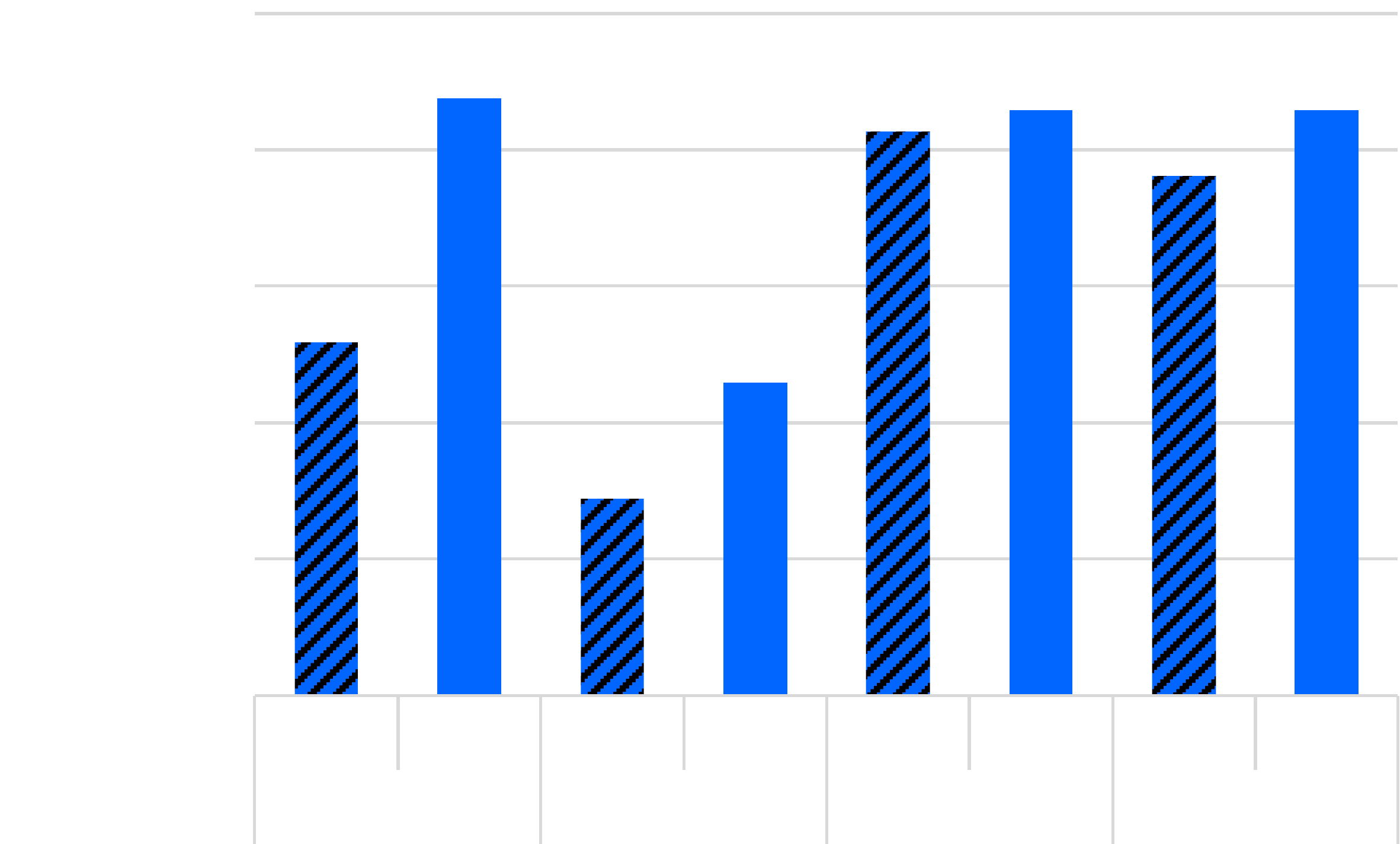}
    \includeinkscape[width=.49\textwidth, pretex=\relscale{0.5}]{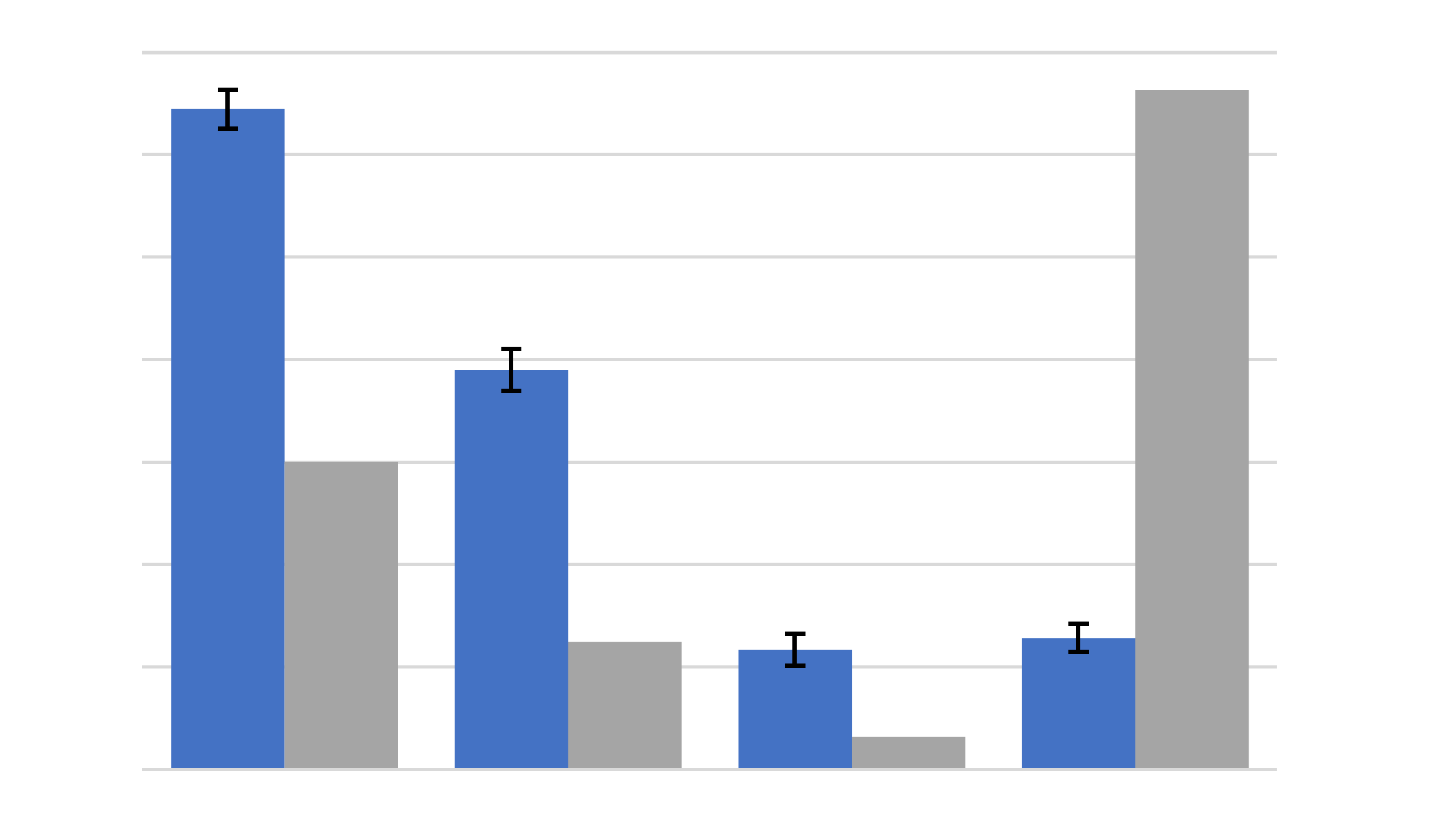}
    \caption{Impact of the fused optimizer on training performance of a DCGAN model. We trained a DCGAN model on the lsun-bedroom dataset, first graph shows the time nedded to complete the training in the different configurations, second graph shows the final accuracy, measured with the inception score and Fréchet inception distance.}
    \label{fused}
\end{figure}

Concerning fused optimizer, we notice a small improvement (2\% to 10\% faster training) when training a fused optimizer as compared to our standard optimizer. The difference appears to be smaller than between classic optimizers, we can suppose that is because the extra-bits values are here accessed separately, whereas they form chunks of 32bits in our base optimizer. The results displayed in \autoref{fused} show that our solution does not perform better than fused-fp32 when using 8 extra-bits, however in this case bf16 training is slower than full precision training (likely because the input data is in fp32 format).

\begin{figure}
    \centering
    \includeinkscape[width=.49\columnwidth, pretex=\relscale{0.5}]{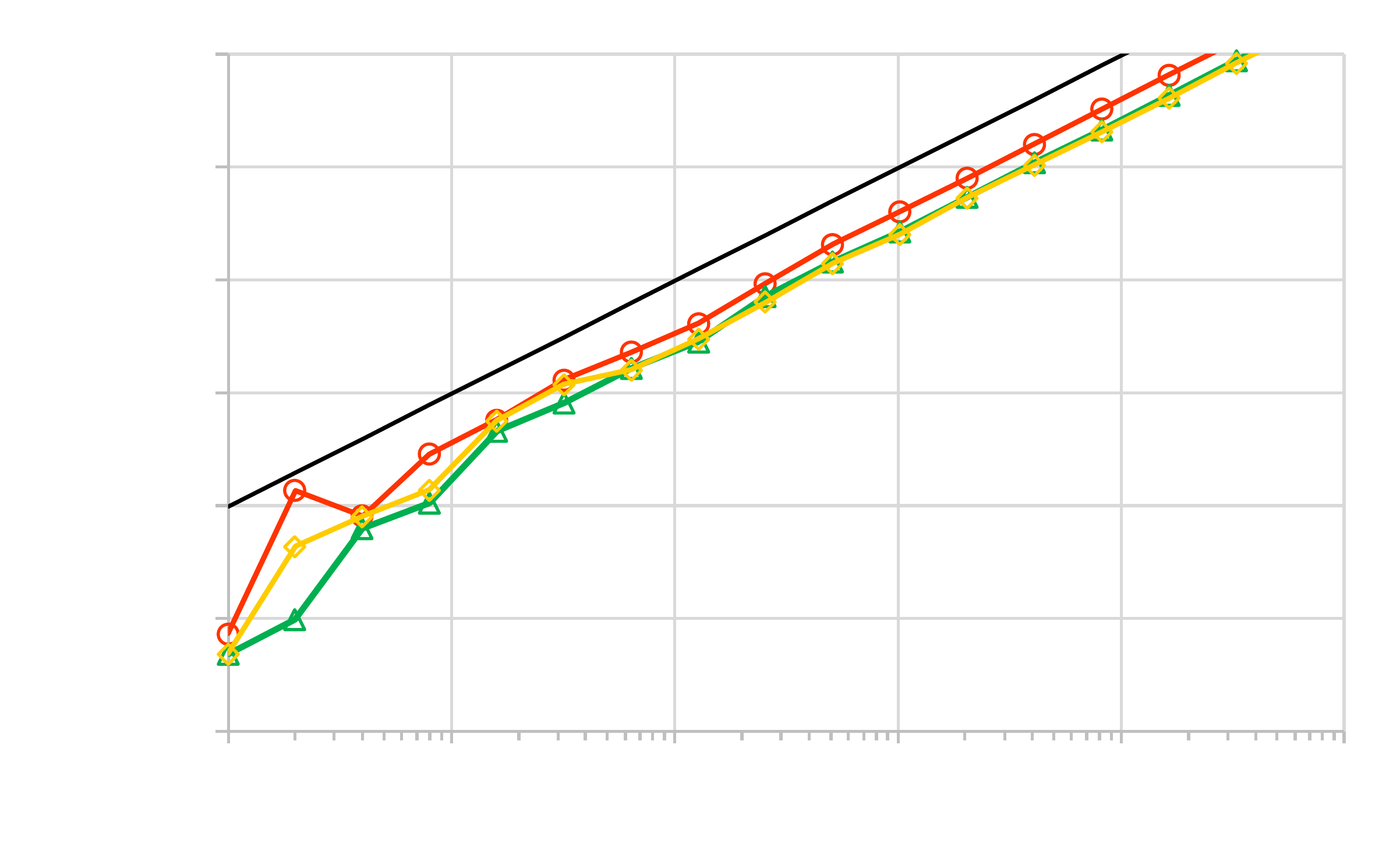}
    \includeinkscape[width=.5\columnwidth, pretex=\relscale{0.5}]{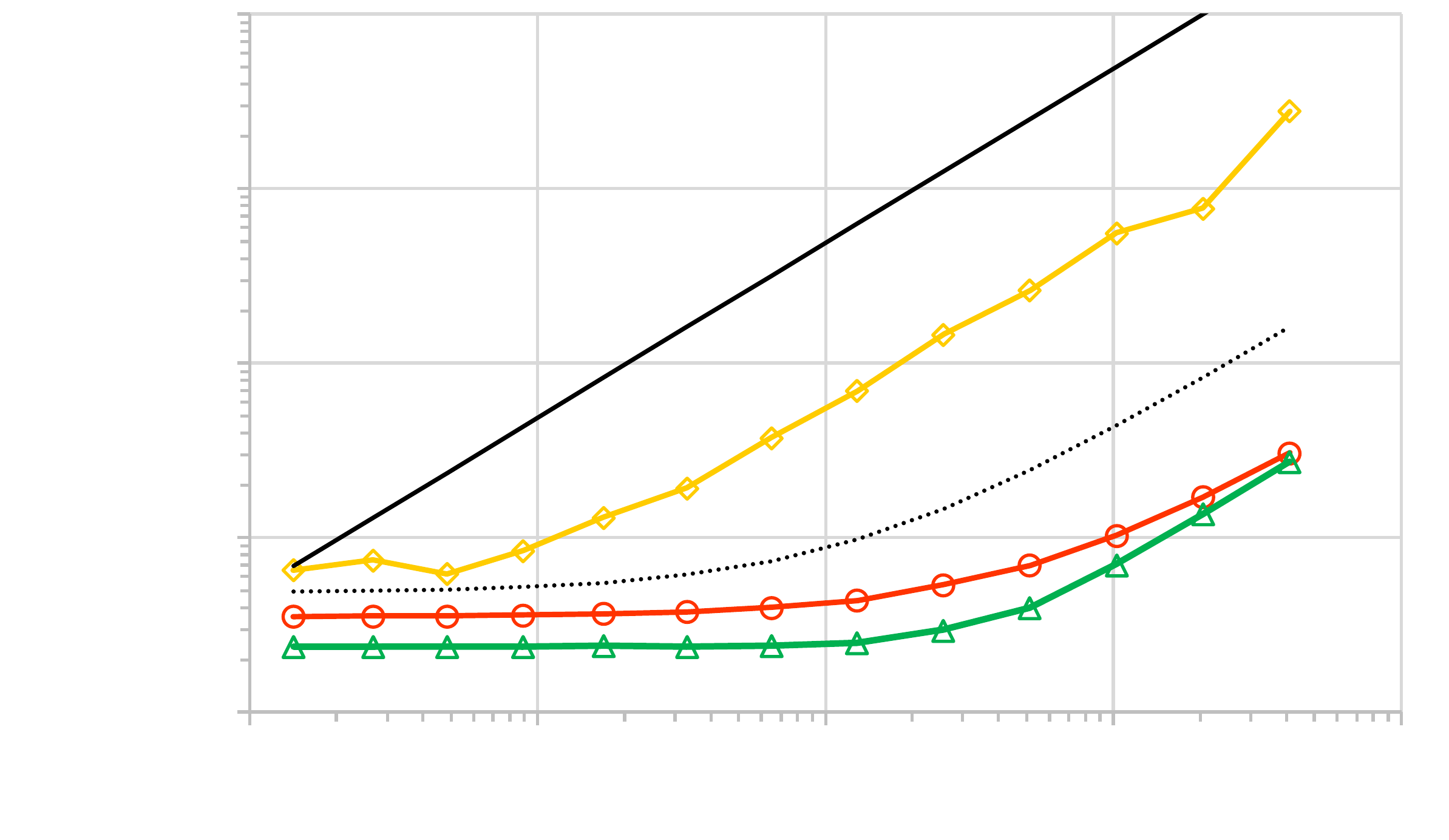}
    \caption{Forward Error On 16bits Tensor Addition. We measure the round-off error of our different half float formats on operations with random values following the standard normal distribution. First graph shows the error after only one operation (in-place add) given some tensor size $n$. Second graph shows the error after several operations, given their condition number. We denote $\varepsilon$ the fp16 round-off precision and $u$ the precision when using 8 extra-bits. We observe on the first operation that basic fp16+8 behaves somewhat poorer than classical fp16, whereas both extra-bit formats produce significantly better results on cumulative operations.  }
    \label{rstoc}
\end{figure}

\paragraph{Stochastic Rounding} Stochastic rounding provides a better accuracy as compared to classic 16bits formats or basic extra-bit format (see \autoref{rstoc}). However, stochastic rounding is less efficient due to the additional operations required to round and `unround` the values, thus leading to performance closer to fp32 training. Moreover, the gain in accuracy compared to the standard formats is preeminent when few operations have been done, which is where the error due to the data is maximal.

\subsection{Limitations} \label{limitations}
\paragraph{Gradient Accumulation} As stated earlier, fusing the backward pass and the optimizer step prevents the use of the gradients after the backward pass. The use of gradient accumulation (stepping the optimizer after several back-propagation) therefore becomes impossible. However, the main point of gradient accumulation is to reduce the need on GPU memory by splitting the batches into smaller mini-batches, and our solution allows for larger batches by getting rid of the gradient, thus reducing the need for such a mechanism.
\paragraph{Gradient Clipping} Gradient clipping is a method designed to prevent exploding gradients by setting a threshold on their values. It is usually applied between the backward pass and the optimizer step, it then becomes impossible when both are fused. One simple solution to mitigate this issue is to perform the clipping through a parameter hook, for instance: 
\begin{small}
\begin{verbatim} 
for p in model.parameters():
    p.register_hook(lambda grad: torch.clamp(grad, -clip_value, clip_value))
\end{verbatim}
\end{small}
The same applies to loss scaling.
\paragraph{Closure} Some training frameworks make use of a "closure" function, which is called during the optimizer step. It is for instance the case with the L-BFGS optimizer, that uses several call to the loss function during the step process. In this case, fusing the backward pass and the stepping may have unexpected consequences on the optimizer behavior, since gradients are not available where they should be. Although such frameworks are -- to our knowledge -- rather uncommon, we warn that stepping the optimizer after each gradient computation might overall trigger issues with some training configuration that we are not aware of.  

\todo[inline]{Discuss limitations: Gradient accumulation, gradient clipping, distributed training (also do experiments)}

\section{Conclusion}
Mixed precision training is an efficient technique to accelerate neural networks training without massive loss in accuracy. We have shown that its memory consumption can be further reduced by tweaking the optimization process. In many cases, the full precision copy can be dispensed with only part of it. Moreover, fusing the backward-pass and the optimizer step reduces peak memory usage by removing the pressure issued by gradients.

Further work could be focused on extending these mechanisms with smaller floating point arithmetic, like 8bits models. We could also investigate their integration in parallel and distributed training framework, as well as less in common optimization setup.
\clearpage

\bibliography{main}
\bibliographystyle{iclr2023_conference}

\clearpage
\appendix
\section{Appendix}
In this section, we detail the frameworks used in our experiments. All experiments were run using Python 3.9, CUDA 11.8 and Pytorch 2.0.1 or above
\paragraph{ResNet18 \& CIFAR10 \protect\footnote{\url{https://github.com/huggingface/pytorch-image-models}}}
We trained a ResNet18 model on the CIFAR10 dataset, results are listed in \ref{metrics}. We did not make use of loss scaling for the experiments as it has not proved necessary. We also kept all modules, including Batch Norm, in fp16 for the same reason. Further hyper parameters are listed below. 

\begin{table}[h]
    \centering
    \begin{tabular}{l c}
         Batch size: & 256 \\
         Epochs: & 200 \\
         Optimizer: &  SGD with momentum \\
         Learning rate: & 0.3 \\
         Warm-up learning rate: & 1e-6\\
         Minimum learning rate: & 1e-5\\
         Weight Decay: & 2e-4 \\
         Scheduler: & cosine (one cycle)\\
         Seed: & \verb|0xB0B| \\
         Warm-up epochs: & 5  
    \end{tabular}
\end{table}

\paragraph{GLUE\protect\footnote{\url{https://github.com/huggingface/transformers/tree/main/examples/pytorch/text-classification}}}
We fine-tuned the bert-base model using the GLUE benchmark. The original weights are the ones provided with standard huggingface transformers package. 
\begin{table}[h]
    \centering
    \begin{tabular}{l c}
         Batch size: & 16 \\
         Epochs: & 3 \\
         Optimizer: &  AdamW \\
         Learning rate: & 2e-5 \\
         Maximum sequence length: & 128\\
         Seed: & \verb|0xB0B| \\
\end{tabular}\end{table}

\paragraph{DLRM\protect\footnote{\url{https://github.com/facebookresearch/dlrm}}}
We used data from the Kaggle-Criteo dataset \footnote{\url{https://ailab.criteo.com/ressources/}}. For the purposes of the demonstration, we did not use any scaling on the gradients during the fp16 trainings. Performance is evalued through the Area Under the Receiver Operating Characteristic Curve (ROC-AUC)\footnote{\url{www.ncbi.nlm.nih.gov/pubmed/2668680}} on a binary case.

\begin{table}[h]
    \centering
    \begin{tabular}{l c}
Sparse feature size: & 16 \\
Bottom MLP: & 13-512-256-64-16 \\
Top MLP: & 512-256-1 \\
Optimizer: & SGD \\
Batchsize: & 128 \\
Training epochs: & 1 \\
Learning rate: & 0.1 \\
Loss function: & BCE \\ 
Seed: & 2023
\end{tabular}\end{table}

\paragraph{DCGAN\protect\footnote{\url{https://github.com/NVIDIA/apex/tree/master/examples/dcgan}}} We used data from the LSUN dataset \footnote{\url{www.yf.io/p/lsun}}. We used bf16 in order to avoid needing loss-scaling. We used the package pytorch-gan-metrics\footnote{\url{https://github.com/w86763777/pytorch-gan-metrics}} to evaluate our model. Evaluation was completed on a set of 50 000 samples.

\begin{table}[h]
    \centering
    \begin{tabular}{l c}
Loss function: & BCE with logistic \\
Optimizer: & Adam \\
Learning rate: & $7.10^{-5}$ \\
$\beta_1$ (first order momentum): & 0.64 \\ 
Batchsize: & 128 \\
Image size: & 64 \\
Latent vector length (nz): & 100 \\
Training epochs: & 10 \\
Seed: & \verb|0xC0FFEE|
\end{tabular}\end{table}

\end{document}